\documentclass[lettersize,journal]{IEEEtran}
\usepackage{amsmath,amsfonts}
\usepackage{algorithmic}
\usepackage{algorithm}
\usepackage{array}
\usepackage[caption=false,font=normalsize,labelfont=sf,textfont=sf]{subfig}
\usepackage{textcomp}
\usepackage{stfloats}
\usepackage{url}
\usepackage{verbatim}
\usepackage{hyperref}
\usepackage{graphicx}
\usepackage{cite}
\usepackage{enumitem}
\usepackage{adjustbox}
\usepackage{booktabs}       
\usepackage{amssymb}    
\usepackage{pifont}
\usepackage{float}
\usepackage{multirow} 
\usepackage{tabularx}  
\usepackage{orcidlink}
\newcolumntype{Y}{>{\centering\arraybackslash}X}
\usepackage{xcolor} 
\colorlet{mycitecolor}{blue!90!black}
\definecolor{hotpink}{RGB}{255, 20, 147}

\hypersetup{
	colorlinks=true,
	citecolor=mycitecolor,
	linkcolor=red,
	urlcolor=black
}
\begin{document}
	\title{DOD-SA: Infrared-Visible Decoupled Object Detection with \\Single-Modality Annotations}
	\author{Hang Jin$^{\orcidlink{0009-0002-0143-2772}}$, Chenqiang Gao$^{\orcidlink{0000-0003-4174-4148}}$, \textit{Member}, \textit{IEEE}, Junjie Guo$^{\orcidlink{0009-0005-8957-8419}}$, Fangcen Liu$^{\orcidlink{0000-0002-1845-9907}}$, Qinyao Chang$^{\orcidlink{0009-0008-7690-6788}}$,\\Kanghui Tian$^{\orcidlink{0009-0001-1006-8493}}$, and Deyu Meng$^{\orcidlink{0000-0002-1294-8283}}$, \textit{Member}, \textit{IEEE}
	\thanks{(\textit{Corresponding author: Chenqiang Gao.})}
	\thanks{Hang Jin, Kanghui Tian and Chenqiang Gao are with the School of Intelligent Systems Engineering, Shenzhen Campus of Sun Yat-sen University, Shenzhen, Guangdong 518107, P.R. China (e-mail: jinx55@mail2.sysu.edu.cn, gaochq6@mail.sysu.edu.cn, tiankh@mail2.sysu.edu.cn).}
	\thanks{Junjie Guo is with the School of Communications and Information Engineering, Chongqing University of Posts and Telecommunications, Chongqing 400065, China (e-mail: s220101036@stu.cqupt.edu.cn).}
	\thanks{Fangcen Liu is with the School of Computer Science and Technology, Hainan University, Haikou 570228, China (e-mail: liufc67@gmail.com).}
	\thanks{Qinyao Chang is with the School of Information and Communication Engineering, University of Electronic Science and Technology of China, Chengdu 611731, China (e-mail: 202422010405@std.uestc.edu.cn).}
	\thanks{Deyu Meng is with the Macau Institute of Systems Engineering, Macau University of Science and Technology, Taipa, Macau, and also with the School of Mathematics and Statistics, Xi’an Jiaotong University, Xi’an 710049, China. (e-mail: dymeng@mail.xjtu.edu.cn).}
	}

\markboth{IEEE Transactions on Geoscience and Remote Sensing,~Vol.~XX, No.~XX, Month~Year}{Author \MakeLowercase{\textit{et al.}}: Title}

\IEEEaftertitletext{}
\maketitle

\begin{abstract}
Infrared-visible object detection has shown great potential in real-world applications, enabling robust all-day perception by leveraging the complementary information of infrared and visible images. However, existing methods typically require dual-modality annotations to output decoupled detection results, leading to high annotation costs and limiting scalability in large-scale remote sensing applications. To address this challenge, we propose a novel infrared-visible \textbf{D}ecoupled \textbf{O}bject \textbf{D}etection framework with \textbf{S}ingle-modality \textbf{A}nnotations, called DOD-SA. It is built upon a Single- and Dual-Modality Collaborative Teacher-Student Network (CoSD-TSNet), which consists of a single-modality branch (SM-Branch) and a dual-modality decoupled branch (DMD-Branch). This design enables cross-modality knowledge transfer from the labeled modality to the unlabeled modality, and facilitates effective cross-branch supervision. To further improve the quality of pseudo-labels, we introduce a Progressive and Self-Tuning Training Strategy (PaST) that trains the model in three stages: 1) SM-Branch self-training, 2) SM-Branch guiding the learning of DMD-Branch, and 3) DMD-Branch refinement. In addition, we design a Pseudo Label Assigner (PLA) to match labels across modalities, explicitly addressing modality misalignment during training. Extensive experiments on four public datasets demonstrate that our method achieves state-of-the-art performance. Our code and datasets have been released at \url{https://github.com/Hang-J/DOD-SA}.
\end{abstract}


\section{Introduction}
\label{sec:intro}
\begin{figure}[t]
	\centering
	\includegraphics[width=1.0\linewidth]{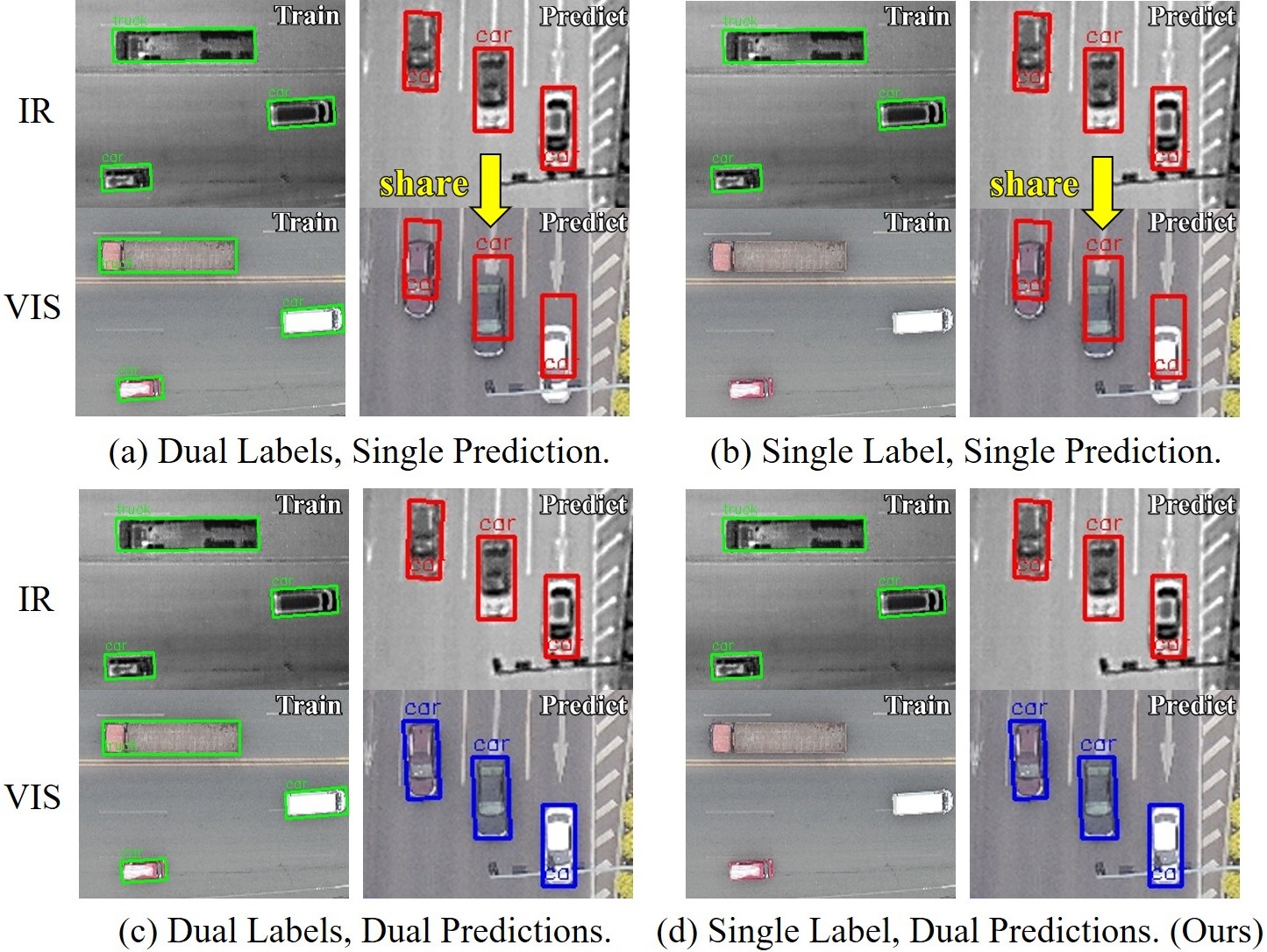}
	\caption{Comparison of multispectral object detection methods. IR and VIS denote the infrared and visible modalities, respectively. Green boxes: ground truth; red/blue boxes: predictions. Our method achieves decoupled object detection in inference by using single-modality training annotations.}
	\label{fig:different}
\end{figure}

\IEEEPARstart{I}{nfrared-visible} object detection\cite{TSFADet,CAGTDet,DDCINet,CALNet,OAFA} is widely used in many fields, including UAV remote sensing and security surveillance, due to its ability to leverage complementary spectral information for robust all‑day perception. However, in practice, infrared (IR) and visible (VIS) images are inevitably misaligned due to differences in imaging time, object motion, or sensor parallax \cite{MFPT}, even if they have been manually pre-registered through operations such as image cropping \cite{Zeng_Li_Cao_Zhang_2022} and affine transformation \cite{Zhang_Ding_2022}. This misalignment disrupts the feature learning of the two modalities, thus affecting the detection performance of the model.

To address this issue, some approaches \cite{zhang2019weakly,TSFADet,CAGTDet} utilized labels from both modalities to explicitly learn cross-modality offsets at the image or feature level, facilitating information fusion. Meanwhile, others \cite{C2former,M2FP,OAFA,guo2024damsdet} enabled the model to implicitly learn such offsets by leveraging single-modality annotations. Nevertheless, these approaches share a single set of bounding box predictions across both modalities, as illustrated in Fig.~\ref{fig:different}(a) and (b).  

In practical applications, object detection typically serves as a preliminary step for downstream processing. To ensure task reliability, modern systems cannot rely solely on single-modality or shared detection results. For example, when tracking targets in multispectral scenarios, reliance on only one modality's detections may cause tracking drift or target loss if the target is poorly localized or invisible in that modality. In such cases, incorporating complementary results from the other modality can effectively mitigate these failures. Therefore, outputting decoupled object positions for each modality is essential, particularly for multi-modal object tracking, complex behavioral analysis, and autonomous navigation of unmanned aerial vehicles (UAVs). Recently, DPDETR \cite{DPDETR} was proposed to address the decoupling problem. As shown in Fig.~\ref{fig:different}(c), this model can output positions of each object in both modalities during the prediction. However, this method requires annotations from both modalities, and the procedure of manually pairing boxes across modalities significantly increases annotation cost and complexity. This raises a pivotal question: Is it possible to attain the benefits of decoupled detection without the substantial overhead of dual-modality labeling? This problem setting encounters two primary challenges. First, there exists inherent spectral discrepancy and spatial misalignments between infrared and visible images. Second, traditional semi-supervised learning frameworks typically assume a shared label space between pseudo-labels and ground truth, thereby failing to account for the coordinate-level divergence essential for decoupled multispectral localization.

Motivated by this, we propose a novel infrared-visible Decoupled Object Detection framework with Single-modality Annotations (DOD-SA). As shown in Fig.~\ref{fig:different}(d), DOD-SA enables decoupled prediction while requiring labels from only one modality during training. Our key idea is to leverage the teacher-student paradigm\cite{sohn2020simple,liu2021unbiased,xu2021end,xu2023efficient} to compensate for the missing labels of the unlabeled modality via pseudo labeling, then to explicitly pair and correct cross-modality boxes so that supervision remains valid under misalignment. Building on this, we progressively train the model from a single-modality detector to a dual-modality decoupled detector, which improves pseudo-label quality and stabilizes the decoupling capability of the model. Specifically, we design a Single- and Dual-Modality Collaborative Teacher-Student Network (CoSD-TSNet) comprising a single-modality branch (SM-Branch) and a dual-modality decoupled branch (DMD-Branch) to address the issue of missing modality annotations by facilitating robust collaborative learning and pseudo labeling. To ensure effective cross-modality knowledge transfer and cross-branch supervision, we propose a Progressive and Self-Tuning Training Strategy (PaST) which includes three stages by initially training the SM-Branch, then guiding the learning of the DMD-Branch by the SM-Branch, and finally refining the DMD-Branch through self-boosted learning. To tackle the label space divergence, we design a dedicated Pseudo Label Assigner (PLA). This module employs shape-aware heuristics with a search region to match pseudo-labels from the unlabeled modality with ground-truth labels from the labeled modality, and maintains a dynamic label pair bag across training epochs, ensuring reliable supervision for decoupled detection. Extensive experiments on four public infrared-visible datasets (DroneVehicle, FLIR, CVC-14, and KAIST) validate the effectiveness and robustness of our proposed method, which is even comparable to fully supervised dual-modality methods. In summary, our contributions can be summarized as follows:

\begin{itemize}
	\item To the best of our knowledge, we are the first to propose a novel framework, termed DOD-SA, which achieves accurate decoupled object detection on both infrared and visible modalities using single-modality annotations.
	\item We design a Single- and Dual-Modality Collaborative Teacher-Student Network that leverages the teacher to generate pseudo-labels for the unlabeled modality. This framework also enables the single-modality branch to guide the dual-modality decoupled branch within a Progressive and Self-Tuning Training Strategy, further improving the model's decoupled detection performance.
	\item We introduce a Pseudo Label Assigner to obtain high-quality label pairs by matching pseudo-labels from unlabeled modality with ground truth from labeled modality, which addresses modality misalignment~explicitly.
	\item We conduct extensive experiments across four datasets: DroneVehicle, FLIR, CVC-14, and KAIST, where DOD-SA consistently establishes SOTA performance.
\end{itemize}

\section{Related Work}

\subsection{Infrared-Visible Object Detection under Weak Misalignment Conditions}

To address the modality misalignment issue, existing methods mainly focus on learning cross-modality offsets, which can be divided into two categories: paired-annotation supervision and single-annotation supervision, depending on whether both modalities are annotated.

\textbf{Paired-Annotation Supervision}. AR-CNN \cite{zhang2019weakly} first tackled modality misalignment by explicitly learning geometric offsets using paired annotations. TSFADet \cite{TSFADet} introduced a region alignment module to predict the scale and angle offsets of the reference proposals. CAGTDet \cite{CAGTDet} further considered the angle offset to make the alignment more accurate. Recently, DPDETR \cite{DPDETR} introduced modality-specific queries and cross-modal attention to achieve decoupled position prediction for both modalities. However, these methods require accurate paired annotations across modalities, which incurs substantial labeling cost and limits scalability in practice.

\textbf{Single-Annotation Supervision}. C$^2$Former \cite{C2former} calculated the attention values between feature points in the reference modality and the other modality to fuse misaligned target features, while OAFA~\cite{OAFA} learned object-aware adaptive feature alignment for weakly misaligned UAV-based multimodal detection. DDCINet~\cite{DDCINet} mitigated modality inconsistency and redundancy through dual-dynamic cross-modal interaction for multimodal remote sensing object detection. MGSANet~\cite{MGSANet} employed multiscale graph representations and graph attention networks to model cross-modal spatial deviations, thereby aligning RGB and thermal features in a latent feature space. DeformCAT \cite{DeformCAT} employed a deformable cross-attention mechanism to capture spatial and geometric correlations. Zhang~et~al. \cite{zhang2025mitigating} introduced a Mean teacher-based box correction module as a subtask to learn more informative representations for both modalities, which also explicitly produces sensed annotations during training. These single-annotation methods still assume shared localization results between modalities, which cannot satisfy applications requiring modality-specific predictions. In contrast, our method enables dual-modality decoupled prediction with single-modality labels, creating a new practical setting.

\subsection{Semi-Supervised Object Detection Method (SSOD)}

SSOD methods aim to reduce annotation cost by utilizing unlabeled data, typically via consistency regularization or pseudo-labeling, when the latter is more widely adopted. STAC \cite{sohn2020simple}  leveraged strong data augmentations to enforce consistency with offline pseudo-labels. Unbiased Teacher \cite{liu2021unbiased} introduced focal loss \cite{lin2017focal} and the Exponential Moving Average (EMA) to address the class imbalance problem and improve pseudo-labels quality. Soft Teacher \cite{xu2021end} used the score of the pseudo-label as the weight of the loss. Efficient Teacher \cite{xu2023efficient} divided the pseudo-labels into certain and uncertain labels. Despite significant advances, existing SSOD methods almost focus solely on single-modality object detection. In this paper, we apply teacher-student networks to dual-modality object detection to address the problem of single-modality annotations.

\subsection{Label Assignment Strategy}

Classical label assignment methods such as ATSS \cite{ATSS}, PAA \cite{PAA}, AutoAssign \cite{AutoAssign}, and TaskAlignedAssigner \cite{Tood} have demonstrated strong performance under fully supervised settings. However, recent studies \cite{chen2022label,liu2022unbiased} revealed that directly applying these strategies to semi-supervised object detection may lead to performance degradation, primarily due to the presence of noisy pseudo-labels and modality inconsistencies. In this paper, we propose a Pseudo Label Assigner to enable application to dual-modality object detection under the single-modality annotations condition. 
\section{Method}
\subsection{Overall Architecture}

As shown in Fig. \ref{fig:3stage}, our DOD-SA is based on the Single- and Dual-Modality Collaborative Teacher-Student Network (CoSD-TSNet), designed to effectively address the issue of the missing annotations. The training process follows Progressive and Self-Tuning Training Strategy (PaST) which consists of three progressive stages. In each stage, specific branches are activated with distinct loss functions to accomplish specific objectives. As training progresses, the quality of pseudo-labels is continuously enhanced. Moreover, we design a Pseudo Label Assigner (PLA) to generate decoupled label pairs in Stage 2 and 3. Additionally, our method is universally effective, regardless of whether the training set contains only visible or only infrared annotations. In this section, we consider the infrared modality as labeled and the visible modality as unlabeled. Further details are provided below.

\subsection{Single- and Dual-Modality Collaborative Teacher-Student Network (CoSD-TSNet)}

\begin{figure*}[htbp]
\centering
\includegraphics[width=1.0\linewidth]{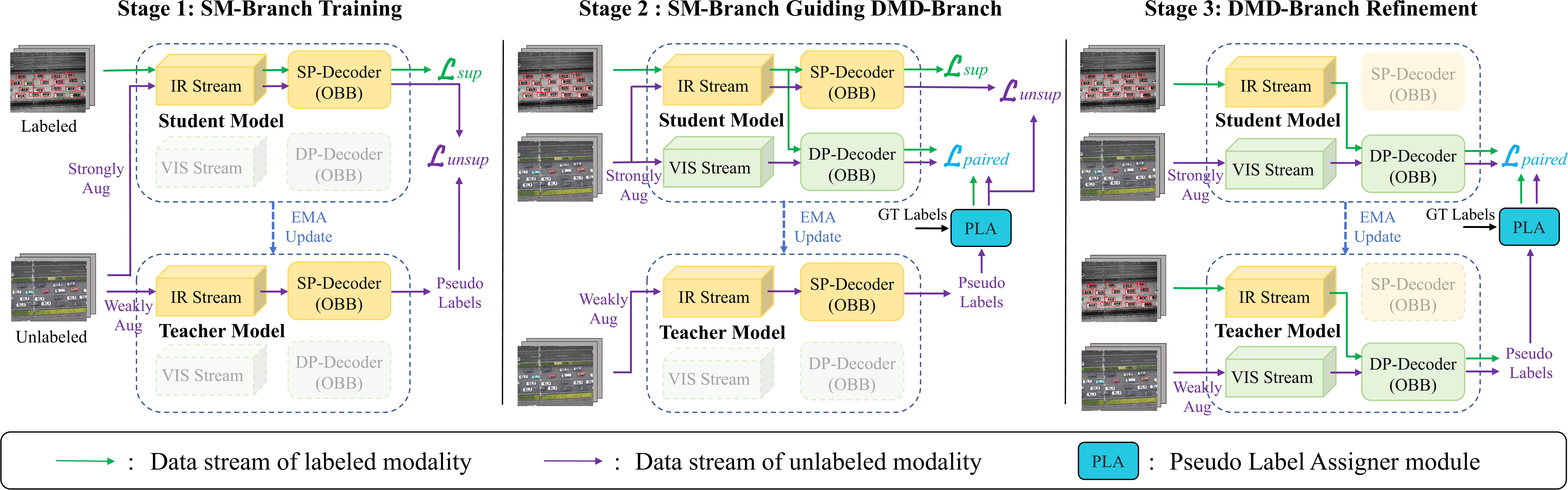}

\caption{The overall framework of our proposed DOD-SA. Our model is based on the CoSD-TSNet to address missing modality annotations through pseudo-label mining. The training follows a three-stage PaST strategy: (1) Stage 1 focuses on SM-Branch (IR Stream and SP-Decoder) self-training; (2) Stage 2 involves the SM-Branch guiding the training of the DMD-Branch (IR Stream, VIS Stream, and DP-Decoder); and (3) Stage 3 performs DMD-Branch refinement via collaborative learning. The PLA module is integrated to match labeled-modality ground truth with unlabeled-modality pseudo-labels. The transparent parts indicate that they are excluded from the computation but remain in the model. For inference, only the teacher model's DMD-Branch is retained. ``OBB'' denotes oriented bounding box.}

\label{fig:3stage}
\vspace{-1em}
\end{figure*}

As presented in Fig.~\ref{fig:3stage}, our model is built upon a teacher-student framework, where the teacher/student model consists of two parameter-unshared streams (IR Stream and VIS Stream), a single-modality decoder (SP-Decoder) and a dual-modality decoupled position decoder (DP-Decoder). Specifically, the IR/VIS streams comprise a backbone, with the possible addition of an encoder, for feature extraction. The SP-Decoder contains a decoder and a detection head, while the DP-Decoder further includes a feature fusion module before these components. For the decoding process, the SP-Decoder processes single-modality features, whereas the DP-Decoder integrates features from both modalities. Moreover, the branch that passes through SP-Decoder is called the single-modality branch (SM-Branch) and the branch that passes through DP-Decoder is called the dual-modality decoupled branch (DMD-Branch). 

For the SM-Branch, it is similar to a single-modality detector, where it can only input an image from a single modality $x_{m}\in\mathbb{R}^{H \times W \times 3}, m\in\{\mathrm{vis},\mathrm{ir}\}$ at a time and outputs one classification result $\hat{c}$ and one modality's position result~$\hat{b}_m$, where $H \times W$ represents the spatial resolution. For the DMD-Branch, it functions as a decoupled position dual-modality detector, where it simultaneously processes images from two modalities $\{x_{ir},x_{{vis}}\}$ and outputs one classification result $\hat{c}$ and decoupled position results for both modalities $\{\hat{b}_{ir},\hat{b}_{{vis}}\}$.

In our framework, the teacher model is responsible for mining pseudo-labels for unlabeled modality. Through the Exponential Moving Average (EMA) update mechanism, the results of the teacher are improved as the student is trained. We design this single- and dual-modality collaborative structure with the idea of using the single-modality detector to guide the training of the dual-modality detector. Additionally, it facilitates knowledge transfer from the labeled modality to the unlabeled modality by effectively leveraging the available labeled data. 

\subsection{Progressive and Self-Tuning Training Strategy (PaST)} \label{three-stage training}
To ensure effective collaborative learning and gradually mine high-quality pseudo-labels of the unlabeled modality, we propose a Progressive and Self-Tuning Training Strategy. As illustrated in Fig.~\ref{fig:3stage}, PaST comprises three stages: (1) Stage~1. This stage aims to initially train the SM-Branch to develop preliminary capabilities for generating visible pseudo-labels. (2) Stage 2. The SM-Branch guides the training of the DMD-Branch to equip it with certain decoupled detection ability. (3) Stage 3. We optimize the entire DMD-Branch to refine decoupled detection results on the infrared and visible modality. Each stage is explained in detail below.

\textbf{Stage 1.} This stage only activates the SM-Branch and comprises two parts: burn-in and teacher-student mutual learning. The burn-in part aims to train the model on labeled data to establish initial detection capabilities for the labeled modality. During the teacher-student mutual learning part, knowledge is transferred from the labeled modality to the unlabeled modality, thereby enhancing the model's detection performance on the unlabeled modality. 

\textit{Burn-in}: The labeled IR data \smash{$x_{ir}$} are fed into the SM-Branch of the student model to obtain the IR predictions \smash{$\hat{y}^{s}_{ir}=\{\hat{c}, \hat{b}_{ir}\}$}, which is supervised by IR ground truth \smash{${y}_{ir}=\{{c}, {b}_{ir}\}$}:
\begin{equation}
\mathcal{L}_{sup}=\mathcal{L}_{cls}(\hat{c},{c})+\mathcal{L}_{box}(\hat{b}_{ir},{b}_{ir}),
\end{equation}
$\mathcal{L}_{cls}$ is the classification loss and $\mathcal{L}_{box}$ is the bounding box regression loss. This part continues for $\mathcal{K}_{1}$ epochs with the loss function $\mathcal{L}_{sup}$. 

\textit{Teacher-Student Mutual Learning}: The student model's SM-Branch processes strongly augmented unlabeled visible data \smash{$x^{*}_{{vis}}$}, while the teacher model's SM-Branch receives a weakly augmented version \smash{$x^{\prime}_{{vis}}$} of the same data. The teacher model outputs visible pseudo-labels \smash{$\tilde{y}^{s}_{{vis}}=\{\tilde{c}, \tilde{b}_{{vis}}\}$} and the student model outputs visible predictions \smash{$\hat{y}^{s}_{{vis}}=\{\hat{c}, \hat{b}_{{vis}}\}$}. To obtain more reliable and certain pseudo-labels, we filter out pseudo-labels below a batch-adaptive threshold which is explained in our proposed PLA, which will be introduced in \ref{PLA}. Then, we use the filtered pseudo-labels to supervise the visible predictions of the student model:
\begin{equation}
\mathcal{L}_{unsup}=\mathcal{L}_{cls}(\hat{c},\tilde{c})+\mathcal{L}_{box}(\hat{b}_{{vis}},\tilde{b}_{{vis}}),
\end{equation}
This part continues \smash{$\mathcal{K}_{2}$} epochs with the loss function \smash{$\mathcal{L}_{sup}+\lambda\mathcal{L}_{unsup}$}. Here, \smash{$\lambda$} is an adaptive weight which linearly increases from 0 to 1.0 as training progresses.

\textbf{Stage 2.} We retain the activated modules of Stage 1 and additionally activate the DMD-Branch in the student model where we copy the IR Stream parameters into the VIS Stream. The input \smash{$x_{ir}$ and $x^{*}_{{vis}}$} are simultaneously fed into the DMD-Branch of the student model, outputting the decoupled detection results \smash{$\hat{y}^{d}=\{\hat{c}, \hat{b}_{ir},\hat{b}_{{vis}}\}$}. The input \smash{$x^{\prime}_{{vis}}$} is fed into the SM-Branch of the teacher model, outputting visible pseudo-labels \smash{$\tilde{y}^{s}_{{vis}}=\{\tilde{c}, \tilde{b}_{{vis}}\}$}. We use PLA to match IR ground truth \smash{${y}_{ir}=\{{c},{b}_{ir}\}$} and visible pseudo-labels $\tilde{y}^{s}_{{vis}}=\{\tilde{c},\tilde{b}_{{vis}}\}$ to form available supervised labels ${y}^d=\{{c}, {b}_{ir},\tilde{b}_{{vis}}\}$. Then, we leverage the matched labels to supervise the DMD-Branch:
\begin{equation}
\mathcal{L}_{paired}=\mathcal{L}_{cls}(\hat{c},{c})+\mathcal{L}_{boxir}(\hat{b}_{ir},{b}_{ir})\\+\mathcal{L}_{boxvis}(\hat{b}_{{vis}},\tilde{b}_{{vis}})\label{stage2.loss}.
\end{equation}

Additionally, we also adopt the \smash{$\mathcal{L}_{sup}$} and \smash{$\mathcal{L}_{unsup}$} mentioned in Stage 1, except that \smash{$\tilde{b}_{{vis}}$} in \smash{$\mathcal{L}_{unsup}$} is now the visible-modality result of PLA. Stage 2 continues \smash{$\mathcal{K}_{3}$} epochs, and the overall loss function of the stage is \smash{$\mathcal{L}_{sup}+\mathcal{L}_{unsup}+\mathcal{L}_{paired}$}. 

\textbf{Stage 3.} This stage only activates the DMD-Branch in the teacher and student model. The input $x_{ir}$ and $x^{\prime}_{{vis}}$ are simultaneously input into the DMD-Branch of the teacher model, outputting decoupled pseudo-labels \smash{$\tilde{y}^d=\{\tilde{c}, \tilde{b}_{ir},\tilde{b}_{{vis}}\}$}. Similarly, the student model outputs decoupled prediction results $\hat{y}^d=\{\hat{c}, \hat{b}_{ir},\hat{b}_{{vis}}\}$. We extract the visible pseudo-labels $\tilde{y}_{{vis}}=\{\tilde{c},\tilde{b}_{{vis}}\}$ from $\tilde{y}^d$. These are then matched with the IR ground truth ${y}_{ir}=\{{c},{b}_{ir}\}$ via PLA to generate supervision ${y}^d=\{{c}, {b}_{ir},\tilde{b}_{{vis}}\}$ for the student model. The loss function $\mathcal{L}_{paired}$ is similar to Eq. (\ref{stage2.loss}), except that \smash{$\tilde{b}_{{vis}}$} is from DP-Decoder of the teacher model. Stage 3 continues $\mathcal{K}_{4}$ epochs with the loss function $\mathcal{L}_{paired}$.

After the three-stage training of the model is completed, only the teacher network’s DMD-Branch structure is retained, ensuring robust and effective decoupled object detection on infrared and visible modalities.
\begin{figure*}[t]
\centering
\includegraphics[width=1.0\linewidth]{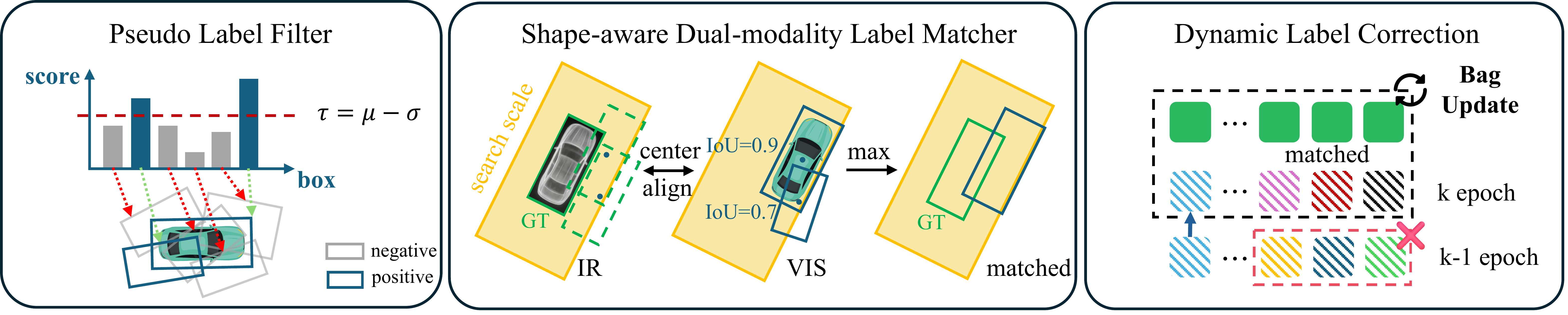}
\caption{Pseudo Label Assigner. It consists of three components: (1) Pseudo Label Filter. It is used to filter out low-quality pseudo-labels. (2) Shape-aware Dual-modality Label Matcher. It matches each infrared ground-truth bounding box with its corresponding visible pseudo label. (3) Dynamic Label Correction. We employ a dynamically updated label pairs bag to continuously correct the pseudo-labels. Solid green blocks: ground truth of the labeled modality. Dashed blocks: pseudo-labels of unlabeled modality. Blue indicates unchanged labels, while other colors represent updated pseudo-labels successfully matched in the current epoch.}
\label{fig:PLA}
\vspace{-1em}
\end{figure*}
\subsection{Pseudo Label Assigner (PLA)}\label{PLA}
To obtain decoupled label pairs and improve the quality of pseudo-labels, we propose the Pseudo Label Assigner. As presented in Fig.~\ref{fig:PLA}, our PLA consists of three submodules: Pseudo Label Filter (PLF), Shape-aware Dual-modality Label Matcher (SDLM) and Dynamic Label Correction (DLC). These submodules are combined to provide DMD-Branch with dual-modality supervision information, through matching the infrared ground truth with visible pseudo-labels. 

\textbf{Pseudo Label Filter (PLF)}. We consider the classification probability of each visible pseudo-label as the score:
\begin{equation}\mathrm{score}_{i}=\max_kp_{i,k}(\tilde{c}),\end{equation}
where $p_{i,k}(\tilde{c})$ denotes the probability of the $i$-th sample being classified as category $k$. Subsequently, similar to \cite{zhang2025mitigating}, we use a batch-adaptive threshold to filter out low-quality samples, which is calculated as follows: 
\begin{equation}
\tau=\mu-\sigma,
\end{equation}
\begin{equation}
\mu=\frac{1}{N}\sum_{i=1}^N\mathrm{score}_i,
\end{equation}
\begin{equation}
\sigma=\sqrt{\frac{1}{N}\sum_{i=1}^N(\mathrm{score}_i-\mu)^2},
\end{equation}
where $\tau$, $\mu$, and $\sigma$ represent the score threshold, the mean, and the standard deviation, respectively, and $N$ denotes the number of samples in a batch. Besides, we initially set a threshold $\tau_{min}$ to filter the pseudo-labels, because there are too many bounding boxes with a low score close to zero, which leads to a distribution of values near zero. These should be ignored first, otherwise they would affect the calculation of the mean.

\textbf{Shape-aware Dual-modality Label Matcher (SDLM)}. After filtering out high-quality visible pseudo-labels, we propose a Shape-aware Dual-modality Label Matcher to pair the set of IR ground-truth bounding boxes $\{b_i^{\mathrm{i r}}\}_{i=1}^N$ with the set of visible pseudo-labels \smash{$\{b_j^{\mathrm{vis}}\}_{j=1}^M$}, where $N$ and $M$ represent their respective quantities. Each bounding box \smash{$b_i=(ct_{i, x},ct_{i, y},w_i,h_i,{\theta_i})$} consists of the center coordinates \smash{$ct_i=(ct_{i, x},ct_{i, y})$}, width $w_i$, height $h_i$, and rotation angle ${\theta_i}$.

For each infrared ground-truth box $b_i^{\mathrm{ir}}$, we construct
a rotated rectangular search region as
\begin{equation}
	\mathcal{S}_i=
	\left\{
	ct_i+
	R(\theta_i)
	\begin{bmatrix}
		\Delta x\\
		\Delta y
	\end{bmatrix}
	\;\middle|\;
	|\Delta x|\leq\frac{\beta w_i}{2},
	\;
	|\Delta y|\leq\frac{\beta h_i}{2}
	\right\}.
\end{equation}
The counterclockwise rotation matrix is defined as
\begin{equation}
	R(\theta_i)=
	\begin{bmatrix}
		\cos\theta_i & -\sin\theta_i\\
		\sin\theta_i & \cos\theta_i
	\end{bmatrix}.
\end{equation}
The hyperparameter $\beta$ controls the scale of the search region.

We use this search region to retain candidates with interior centers while excluding those already matched, obtaining the visible candidate set \smash{$C_i^{\mathrm{vis}}$}:
\begin{equation}
C_i^{\mathrm{vis}}=\left\{b_j^{\mathrm{vis}}: ct_j \in \mathcal{S}_i \text { and } b_j^{\mathrm{vis}} \notin P\right\},
\end{equation}
where $P$ is the set of already paired boxes. We then compute the intersection over union (IoU) between IR ground-truth box and all visible boxes in \smash{$C_i^{\mathrm{vis}}$}, then pair the IR box with the visible candidate having the highest IoU. Since object shapes are nearly identical across modalities, the best-overlapping visible pseudo-label likely corresponds to the same object.
\begin{equation}
b_j^*=\arg \max _{b_j^{\mathrm{vis}} \in C_i^{\mathrm{vis}}} \operatorname{IoU}\left(b_i^{\mathrm{ir}}, b_j^{\mathrm{vis}}\right),
\end{equation}
where $b_j^*$ denotes the successfully matched visible pseudo-label box.

\textbf{Dynamic Label Correction (DLC)}. Considering relying solely on matched label pairs results in insufficient supervision, we incorporate the unmatched cases and maintain a dynamically updated label pair bag to obtain more high-quality label pairs.

In the first epoch of Stage 2, we copy the unmatched infrared ground truth bounding boxes $U_i^{\mathrm{ir}}$ to the visible modality. These are then combined with the matched label pairs $P_0$ to initialize the dual-modality decoupled ground truth $G_0$:
\begin{equation}
	U_0^{\mathrm{ir}}
	=
	\left\{
	b_i^{\mathrm{ir}}
	\;\middle|\;
	\nexists\, b_j^{\mathrm{vis}},
	\left(b_i^{\mathrm{ir}}, b_j^{\mathrm{vis}}\right)\in P_0
	\right\},
\end{equation}
\begin{equation}
	G_0
	=
	P_0
	\cup
	\left\{
	(b,b)
	\;\middle|\;
	b\in U_0^{\mathrm{ir}}
	\right\}.
\end{equation}
In the $k$-th epoch, let $(b_k^{ir}$,$b_k^{vis})$ be one of the label pairs in $G_k$, where $G_k$ denotes the dynamically updated infrared-visible label pairs bag in the $k$-th epoch. If in the next epoch, $b_k^{ir}$ is successfully matched with a new visible pseudo-label $b_k^{vis*}$, this label pair will be updated to \smash{$(b_k^{ir}$,$b_k^{vis*})$}. Otherwise, the label pair remains unchanged.

\section{Experiments}
\subsection{Datasets and Evaluation Metrics}
\textbf{Datasets.} We conduct experiments on four datasets: DroneVehicle, FLIR, CVC-14, and KAIST. In the following section, we will introduce these datasets in detail.

\begin{enumerate}
\item \textit{DroneVehicle}: This dataset\cite{DroneVehicle} is a large-scale UAV remote sensing object detection benchmark containing drone-captured paired infrared and visible imagery. It contains 28,439 IR-VIS image pairs at a resolution of $840 \times 712$. Each modality has its own set of oriented bounding box annotations, covering five vehicle types: car, bus, truck, van, and freight car. To standardize the experimental protocol, we follow the methodology outlined in DPDETR~\cite{DPDETR} to process the annotations. After preprocessing, the dataset comprises 17,990 image pairs for training and 1,469 pairs for test. In each image pair, the visible and infrared labels maintain a strictly one-to-one correspondence.
\item \textit{KAIST}: The original KAIST Multispectral Pedestrian Dataset\cite{KAIST} consists of 95,328 infrared-visible pairs at a resolution of $640 \times 512$ captured in campus and street environments. Since the original dataset had annotation issues, we utilize the processed dataset which contains 8,593 training pairs with corrected infrared-visible annotations \cite{zhang2019weakly} and 2,252 testing pairs with improved infrared annotations \cite{liu2016multispectral}, following established protocols \cite{C2former,DPDETR,DeformCAT}. In the entire test set, each annotation includes illumination, scale, and occlusion information.
\item \textit{CVC-14}: This pedestrian dataset\cite{CVC-14} is constructed from multispectral video sequences captured by a vehicle navigating street environments, with the image resolution of $640 \times 470$. The dataset is officially divided into a training set of 7,085 frames and a test set of 1,433 frames. It comprises ``Day'' and ``Night'' subsets, recording both visible and infrared modalities. Unlike KAIST, the annotations in CVC-14 do not have occlusion information. Following DeformCAT \cite{DeformCAT}, we exclude images lacking valid annotations across both modalities. Consequently, the final curated dataset comprises 1,497 image pairs for training and 1,304 for testing. To obtain paired infrared-visible labels, we keep the visible labels unchanged and obtain the corresponding infrared labels following the same processing  used in the DroneVehicle dataset.
\item \textit{FLIR}: The original FLIR ADAS dataset\cite{FLIR} provides 14,452 RGBT image pairs captured from an on-road driving perspective. It includes annotations for three main object classes: person, car, and bicycle. We utilize the aligned version \cite{zhang2020multispectral} with a resolution of $640 \times 512$, which consists of 4,129 pairs for training and 1,013 pairs for testing. In this version, the majority of image pairs are well-aligned, with a small number of cases remaining misaligned. Notably, only the infrared images in this dataset are provided with annotations.
\end{enumerate}

\begin{figure}[h]
\centering
\includegraphics[width=1.0\linewidth]{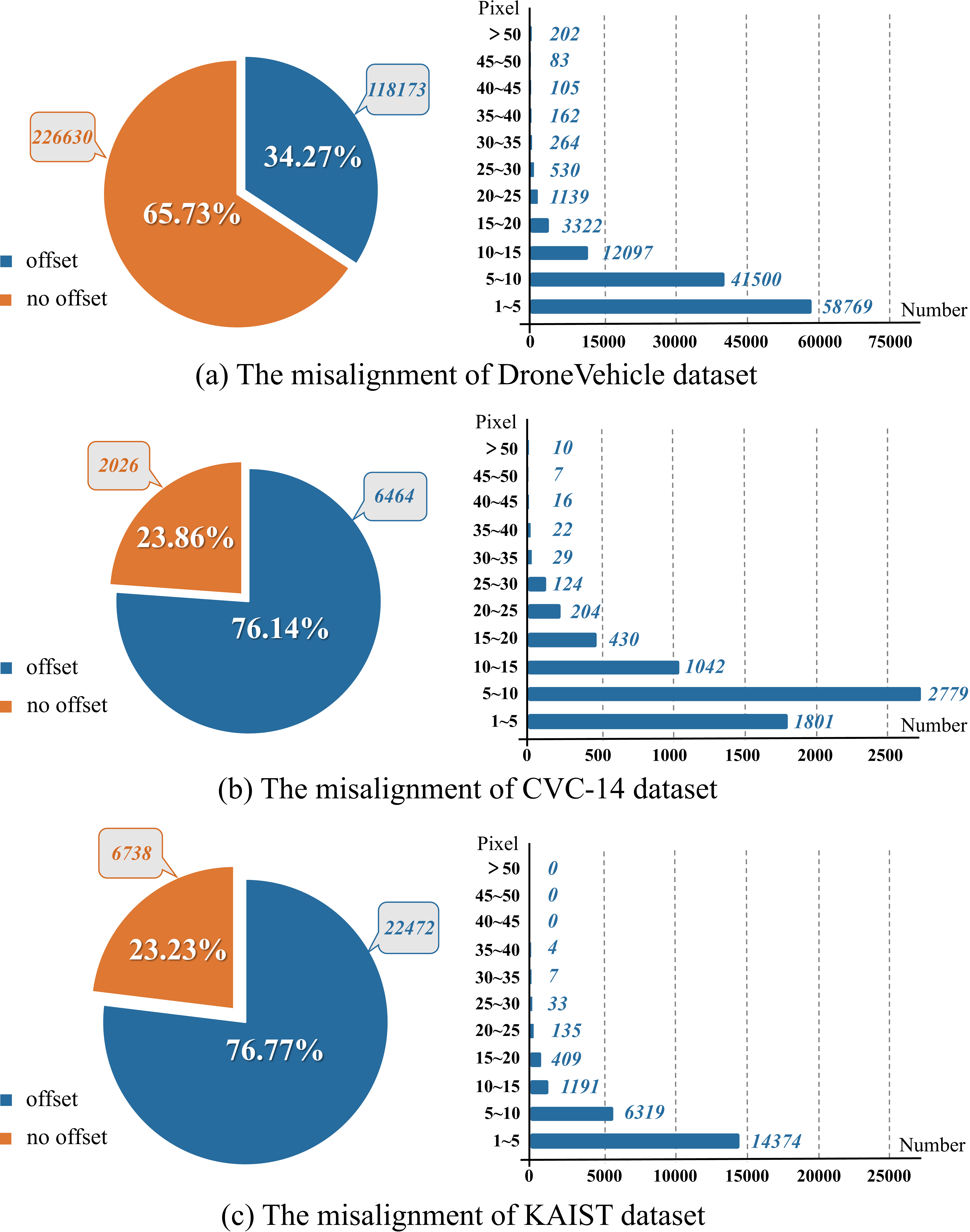}
\caption{We analyze the statistics of misalignment ratios (left) and center pixel shifts (right) between corresponding ground truth bounding boxes in IR-VIS image pairs across the DroneVehicle, CVC-14, and KAIST datasets.}
\label{fig:misalignment}
\end{figure}

Furthermore, we conduct a detailed analysis into the misalignment of the datasets. Since the FLIR dataset only provides annotations for infrared images, quantitative analysis for this dataset is not available. As shown in Fig.~\ref{fig:misalignment}, misalignment is prevalent and severe across the three datasets. They exhibit similar center pixel shift distributions with most values ranging from 1 to 20 pixels, and CVC-14 has the largest average offset. This finding highlights the importance of decoupled detection.

\textbf{Evaluation metrics.} We employ different evaluation metrics tailored to the specific characteristics of each dataset. For the DroneVehicle dataset, we adopt the mean Average Precision at an IoU threshold of 0.5 ($m$AP50) as the metric. For the pedestrian detection benchmarks, we report a more comprehensive set of metrics on the FLIR dataset, including $m$AP50, $m$AP75, and the standard COCO-style $m$AP (averaged over IoU thresholds from 0.5 to 0.95). For KAIST and CVC-14, we utilize the Log-Average Miss Rate (MR$^{-2}$) calculated over the False Positives Per Image (FPPI) range of $[10^{-2}, 10^{0}]$ under the IoU threshold of 0.5. For KAIST, to fully investigate the performance of the detector, we evaluate across three subsets: \textit{Illumination} (Day, Night, and All), \textit{Scale} (Near, Medium, and Far) and \textit{Occlusion} (None, Partial, and Heavy), following the protocol in C$^2$Former\cite{C2former}. For CVC-14, we follow the \textit{Reasonable} (Day, Night, and All) subset setting\cite{DeformCAT}. It should be noted that a higher $m$AP and a lower MR$^{-2}$ value indicate superior performance.
\begin{table}[t]
\centering
\caption{Detection results ($m$AP50, in \%) on the DroneVehicle dataset.}
\resizebox{\columnwidth}{!}{
\fontsize{14}{16}\selectfont
\begin{tabular}{lcccc}
	\toprule
	Method & Train Images & Train Labels & IR-Test & VIS-Test \\
	\midrule 
	RoI Transformer\cite{RoI_Transformer} & IR & IR  & 63.65 & 23.62 \\
	S$^2$ANet\cite{s2adet} & IR & IR  & 71.71 & 38.94 \\
	RT-DETR (OBB) \cite{RTDETR}& IR & IR  & 77.29 & 39.52  \\
	\midrule
	RoI Transformer\cite{RoI_Transformer} & VIS & VIS  & 26.69 & 55.88 \\
	S$^2$ANet\cite{s2adet} & VIS & VIS  & 38.98 & 66.48 \\
	RT-DETR (OBB)\cite{RTDETR} & VIS & VIS & 46.51 & 72.92 \\
	\midrule
	AR-CNN (OBB)\cite{AR-CNN} & IR+VIS & IR+VIS & 72.21 & 69.78 \\
	TSFADet\cite{TSFADet} & IR+VIS & IR+VIS & 73.06 & - \\
	CAGTDet\cite{CAGTDet} & IR+VIS & IR+VIS & 74.57 & - \\
	DPDETR \cite{DPDETR}& IR+VIS & IR+VIS & 79.90 & 79.81 \\	
	\midrule
	C$^2$Former\cite{C2former} & IR+VIS & IR & 60.44 & 58.49 \\
	CIAN (OBB) \cite{CIAN}& IR+VIS & IR & 70.23 & - \\
	MC-DETR \cite{MC-DETR} & IR+VIS & IR & 76.90 & - \\
	DDCINet\cite{DDCINet} & IR+VIS & IR & 78.40 & - \\
	CCLDet\cite{CCLDet} & IR+VIS & IR & 79.06 & 76.87  \\
	\textbf{DOD-SA (Ours)}  & IR+VIS & IR  & \textbf{80.41} & \textbf{78.87} \\
	\midrule
	E2E-MFD\cite{E2E-MFD} & IR+VIS & VIS & 76.17  & 76.58 \\
	CALNet\cite{CALNet} & IR+VIS & VIS  & 76.20  & 77.10 \\
	\textbf{DOD-SA (ours)}  & IR+VIS & VIS  & \textbf{77.96} & \textbf{78.19} \\
	\bottomrule
\end{tabular}
}

\vspace{1ex}
\parbox{0.98\linewidth}{\footnotesize
Single-modality methods are evaluated separately on IR and VIS inputs; shared-output methods use a common prediction set, whereas decoupled methods use modality-specific outputs. The best results within each comparison group are shown in \textbf{bold}, and ``-'' denotes unavailable results.
}
\label{table1}
\end{table}

\begin{table}[t]
\centering
\caption{Detection results ($m$AP50/$m$AP75/$m$AP, in \%) on the FLIR Dataset.}
\resizebox{1.0\columnwidth}{!}{
\fontsize{14}{16}\selectfont
\begin{tabular}{lccccc}
	\toprule
	\multirow{2}{*}[-0.5ex]{Method} & \multirow{2}{*}[-0.5ex]{Train image} & \multirow{2}{*}[-0.5ex]{Train label} & \multicolumn{3}{c}{IR-Test} \\ 
	\cmidrule(lr){4-6}
	& & & $m$AP50 & $m$AP75 & $m$AP \\
	\midrule
	
	YOLOv11 \cite{yolov11} & IR & IR  & 78.8 & - & 42.8\\
	DINO \cite{dino} & IR & IR  & 80.6 & 42.7 & 44.8\\
	DEIMv2 \cite{DEIMv2} & IR & IR  & 80.8 & 39.5 & 43.8\\
	\midrule
	GAFF \cite{GAFF}& IR+VIS & IR & 72.9 & 32.9 & 36.6 \\
	SMPD \cite{SMPD}& IR+VIS & IR & 73.6 & - & -\\
	CFT \cite{CFT} & IR+VIS & IR & 78.7 & 35.5& 40.2 \\
	ICAFusion \cite{icafusion} & IR+VIS & IR & 79.2 & 36.9& 41.4 \\
	MFPT \cite{MFPT}& IR+VIS & IR & 80.0 & -& - \\
	LRAF-Net\cite{lraf} & IR+VIS & IR & 80.5 & -& 42.8 \\
	DeformCAT\cite{DeformCAT} & IR+VIS & IR & 84.7 & 39.3& 43.2 \\
	\textbf{DOD-SA(ours)}  & IR+VIS & IR  & \textbf{85.2} & \textbf{44.5} & \textbf{46.8} \\
	\bottomrule
\end{tabular}
}

\vspace{1ex}
\parbox{0.98\linewidth}{\footnotesize
The best results are highlighted in \textbf{bold}. ``-'' indicates that the results are unavailable.
}
\label{table2}
\end{table}

\subsection{Implementation Details}
We select DPDETR \cite{DPDETR} as our baseline framework due to its superior decoupled detection capabilities. Specifically, each of the IR and VIS streams in our model is composed of a ResNet-50 \cite{he2016deep} backbone and an efficient encoder with a feature level of $L=3$. The SP-Decoder adopts the decoder architecture of the RT-DETR \cite{RTDETR}, while the DP-Decoder follows the decoder of DPDETR. Both decoders consist of six layers. We set the number of attention heads to $H=8$ and the number of object queries to $N=300$. For optimization, we employ the AdamW optimizer with a weight decay of $1\times 10^{-4}$. The learning rate is initialized at $1\times 10^{-7}$, linearly warmed up to $1\times 10^{-4}$, and subsequently reduced to $1\times 10^{-5}$ using a piecewise decay schedule. Our data augmentation strategy consists of weak augmentation (random rotation and flipping) and strong augmentation (photometric transformations, random erasing, and blurring). The batch size is set to 4, and the Exponential Moving Average (EMA) decay weight is 0.9999. For the DroneVehicle dataset, the training epochs are set as $\mathcal{K}_{1}=20$, $\mathcal{K}_{2}=10$, $\mathcal{K}_{3}=15$, and $\mathcal{K}_{4}=20$, with the PLA hyperparameter $\beta$ set to 1.0. For the KAIST, CVC-14, and FLIR datasets, the training epochs are configured as $\mathcal{K}_{1}=6$, $\mathcal{K}_{2}=6$, $\mathcal{K}_{3}=12$, and $\mathcal{K}_{4}=20$, with $\beta$ set to 1.2. The number of epochs is configured to strike a balance between achieving full convergence at every stage and reducing the total training time. All experiments are conducted on a single NVIDIA RTX A6000 GPU.

\begin{table*}[!ht]
\centering
\caption{Detection results (MR$^{-2}$, in \%) on the CVC-14 dataset.}
\label{tab:cvc14_results}
\resizebox{0.7\textwidth}{!}{
\fontsize{8}{10}\selectfont
\setlength{\aboverulesep}{0pt} 
\setlength{\belowrulesep}{2pt}
\begin{tabular}{l c c c c c c c c}
	\toprule
	\multirow{2}{*}[-0.2ex]{Method} & \multirow{2}{*}[-0.2ex]{Train image} & \multirow{2}{*}[-0.2ex]{Train label} & \multicolumn{3}{c}{VIS-Test} & \multicolumn{3}{c}{IR-Test} \\
	\cmidrule(lr){4-6} \cmidrule(lr){7-9} 
	& & & Day & Night & All & Day & Night & All \\
	\midrule
	AR-CNN \cite{AR-CNN}  & IR+VIS  & IR+VIS & 18.7 & 24.3 & 20.5 & 71.8 & 51.4 & 63.5 \\
	DPDETR \cite{DPDETR} & IR+VIS  & IR+VIS & 14.3 & 15.7 & 15.1 & 17.2 & 11.4 & 14.9 \\
	\midrule
	MLPD \cite{MLPD}    & IR+VIS  & VIS & 49.0 & 37.1 & 44.0 & 73.6 & 51.1 & 65.6 \\
	MBNet \cite{MBnet}  & IR+VIS  & VIS & 45.6 & 34.3 & 42.0 & 72.1 & 49.5 & 63.2 \\
	MCHE-CF \cite{mche} & IR+VIS  & VIS & 24.0 & 17.5 & 21.3 & - & - & - \\
	SCDNet \cite{SCDNet}   & IR+VIS  & VIS & 20.3 & 12.9 & 19.0 & - & - & - \\
	DeformCAT \cite{DeformCAT} 	 & IR+VIS  & VIS & 14.6 & 12.3 & 13.4 & 67.3 & 38.2 & 57.0 \\
	\textbf{DOD-SA (ours)} & IR+VIS  & VIS & \textbf{13.5} & \textbf{11.3} & \textbf{12.3} & \textbf{29.0} & \textbf{14.8} & \textbf{23.1} \\
	\bottomrule
\end{tabular}
}

\vspace{1ex}
\parbox{0.66\linewidth}{\footnotesize
The best results are highlighted in \textbf{bold}. ``-'' indicates that the results are unavailable. Day, Night and All belong to ``\textit{Reasonable}'' subset, which considers only pedestrians taller than 55 pixels that are not truncated by image boundaries, regardless of their occlusion level.
}
\label{table3}
\end{table*}

\begin{table*}[!ht]
\centering
\caption{Detection results (MR$^{-2}$, in \%) on the KAIST dataset.}
\label{table4}
\resizebox{1.0\textwidth}{!}{
\fontsize{8}{10}\selectfont
\setlength{\aboverulesep}{0pt} 
\setlength{\belowrulesep}{2pt}
\begin{tabular}{l cc ccc ccc ccc}
	\toprule 
	\multirow{3}{*}[-0.7ex]{Method} & \multirow{3}{*}[-0.7ex]{Train image} & \multirow{3}{*}[-0.7ex]{Train label} & \multicolumn{9}{c}{IR-Test} \\
	\cmidrule(lr){4-12} 
	& & & \multicolumn{3}{c}{\textit{Scale}} & \multicolumn{3}{c}{\textit{Occlusion}} & \multicolumn{3}{c}{\textit{Illumination}} \\
	\cmidrule(lr){4-6} \cmidrule(lr){7-9} \cmidrule(lr){10-12} 
	& & & Near & Medium & Far & None & Partial & Heavy & Day & Night & All \\
	\midrule 
	AR-CNN \cite{AR-CNN}     & IR+VIS & IR+VIS   & 0.00  & 16.08 & 69.00 & 31.40 & 38.63 & 55.73 & 34.36 & 36.12 & 34.95 \\
	TSFADet \cite{TSFADet}     & IR+VIS & IR+VIS   & 0.00  & 15.99 & 50.71 & 25.63 & 37.29 & 65.67 & 31.76 & 27.44 & 30.74 \\	
	CAGTDet \cite{CAGTDet}      & IR+VIS & IR+VIS  & 0.00  & 14.00 & 49.40 & 24.48 & 33.20 & 59.35 & 28.79 & 27.73 & 28.96 \\
	DPDETR \cite{DPDETR}      	& IR+VIS & IR+VIS	& 0.00 & 15.54 & 36.84 & 21.09 & 26.61 & 57.05 & 26.75 & 21.01 & 25.04 \\			
	\midrule
	ACF \cite{KAIST}     & IR+VIS & IR       & 28.74 & 53.67 & 88.20 & 62.94 & 81.40 & 88.08 & 64.31 & 75.06 & 67.74 \\
	Halfway Fusion \cite{Halfway_Fusion}& IR+VIS & IR & 8.13  & 30.34 & 75.70 & 43.13 & 65.21 & 74.36 & 47.58 & 52.35 & 49.18 \\
	FusionRPN+BF \cite{fusionRPN} & IR+VIS & IR  & 0.04  & 30.87 & 88.86 & 47.45 & 56.10 & 72.20 & 52.33 & 51.09 & 51.70 \\
	IAF R-CNN \cite{IAF}   & IR+VIS & IR   & 0.96  & 25.54 & 77.84 & 40.17 & 48.40 & 69.76 & 42.46 & 47.70 & 44.23 \\
	CIAN \cite{CIAN}        & IR+VIS & IR   & 3.71  & 19.04 & 55.82 & 30.31 & 41.57 & 62.48 & 36.02 & 32.38 & 35.53 \\
	MSDS-RCNN \cite{MSDS}   & IR+VIS & IR   & 1.29  & 16.19 & 63.73 & 29.86 & 38.71 & 63.37 & 32.06 & 38.83 & 34.15 \\
	MBNet \cite{MBnet}      & IR+VIS & IR   & 0.00  & 16.07 & 55.99 & 27.74 & 35.43 & 59.14 & 32.37 & 30.95 & 31.87 \\
	CMPD \cite{CMPD}     & IR+VIS & IR   & 0.00  & 12.99 & 51.22 & 24.04 & 33.88 & 59.37 & 28.30 & 30.56 & 28.98 \\
	C$^2$Former \cite{C2former}  & IR+VIS & IR  & 0.00  & 13.71 & 48.14 & 23.91 & 32.84 & 57.81 & 28.48 & 26.67 & 28.39 \\
	DeformCAT \cite{DeformCAT}     & IR+VIS & IR   & \textbf{0.00}  & \textbf{12.31} & 33.61 & \textbf{19.09} & \textbf{22.11} & \textbf{45.48} & \textbf{25.94} & 16.27 & \textbf{22.90} \\
	\textbf{DOD-SA (ours)}& IR+VIS & IR & 0.81 & 14.55 & \textbf{33.28} & 20.17 & 23.50 & 49.31 & 27.79 & \textbf{15.73} & 23.92 \\
	\bottomrule 
\end{tabular}
} 

\vspace{1ex}
\parbox{0.95\linewidth}{\footnotesize
``\textit{Illumination}'' considers all heights and all occlusion levels under different light conditions. ``\textit{Scale}'' is categorized by height into three types: Near ($\ge 115$ pixels), Medium ($[45, 115)$ pixels), and Far ($< 45$ pixels). ``\textit{Occlusion}'' is classified as None (no occlusion), Partial (up to 50\% occluded), and Heavy (more than 50\% occluded). Furthermore, all pedestrians beyond the image boundaries are ignored. The best results are highlighted in \textbf{bold}.
}
\end{table*}

\subsection{Comparisons with state-of-the-art}
We compare our proposed DOD-SA with several recent state-of-the-art methods on several large public datasets, including DroneVehicle, FLIR, CVC-14, and KAIST.

\textbf{Quantitative results for DroneVehicle.} As shown in Table.~\ref{table1}, single-modality methods (RoI Transformer, S$^2$ANet, and RT-DETR) suffer severe performance degradation when models trained on one modality are directly evaluated on the other modality. Certain dual-modality methods (AR-CNN, C$^2$Former, and CCLDet) also lack decoupling capability and thus cannot provide accurate modality-specific localization under misalignment. In contrast, our DOD-SA outperforms other methods under the same training label conditions. Compared to dual-modality methods, it achieves state-of-the-art performance (80.41\% in IR-Test, 78.87\% in VIS-Test) when trained only on IR labels, surpassing CCLDet by 1.4\% $m$AP in IR-Test and 2.0\% in VIS-Test. Even when trained on visible labels, our method delivers strong decoupled results (77.96\% in IR-Test, 78.19\% in VIS-Test), demonstrating effective modality-generalization. Furthermore, our method achieves performance comparable to DPDETR, when DPDETR requires labels from two modalities.

\textbf{Quantitative results for FLIR.} Since the FLIR dataset provides annotations exclusively for IR images, all methods are trained using only IR labels and evaluated on the IR-Test set. The results are shown in Table.~\ref{table2}. Our method achieves SOTA performance under identical conditions, surpassing DeformCAT by 0.5\% in $m$AP50 and showing an even greater advantage in $m$AP. This shows that DOD-SA remains highly effective even when applied to well-aligned datasets like FLIR. This also validates the effectiveness of considering unmatched ground truth labels in our DLC strategy, enabling the model to adapt to zero-offset scenarios from the very beginning.

\textbf{Quantitative results for CVC-14.} Due to the significant misalignment in the CVC-14 dataset, previous methods suffer from severe performance degradation when directly applying visible detection results to the infrared modality as shown in Table.~\ref{table3}. In contrast, our method not only achieves SOTA performance on VIS-Test, but significantly mitigates this degradation on IR-Test. We observe that, similar to the case with the DroneVehicle dataset, DOD-SA performs slightly worse than DPDETR on the unlabeled modality. This is primarily attributed to the difficulty of mining pseudo-labels. 

\textbf{Quantitative results for KAIST.} The KAIST test set only contains infrared annotations\cite{liu2016multispectral}, so the performance metrics for the visible modality cannot be obtained. As evident from Table.~\ref{table4}, although our method is slightly inferior to DeformCAT in terms of IR metrics (with a 1\% higher MR-2 on the `All' subset), it achieves a valuable decoupling capability comparable to DPDETR without requiring labels from two modalities. It surpasses DPDETR on the labeled modality by achieving lower MR-2 scores for almost every item across the three category subsets, which sufficiently demonstrates the effectiveness of our method.

\begin{figure*}[!t]
\centering
\includegraphics[width=0.9\linewidth]{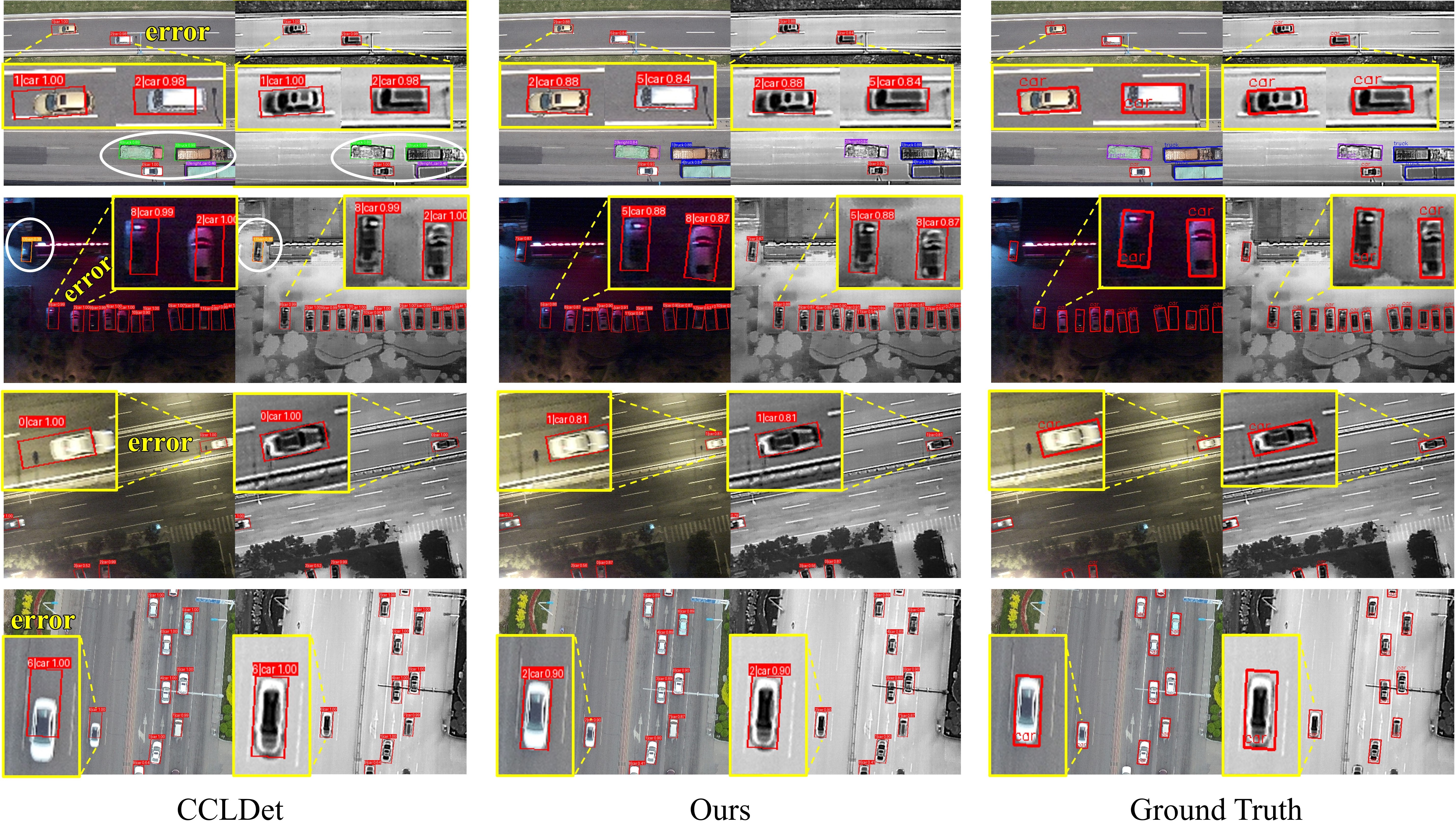}
\caption{Representative results of different methods on the DroneVehicle dataset. For each method, the first column displays visible images, and the second column displays infrared images. The white circles indicate classification errors, and ``error'' denotes mistakes caused by detection offsets. The confidence threshold is set to 0.2 when visualizing these results.}
\label{fig:dronevehicle}
\vspace{-1em}
\end{figure*}

\begin{figure*}[h]
\centering
\vspace*{-\topskip}
\includegraphics[width=0.9\linewidth]{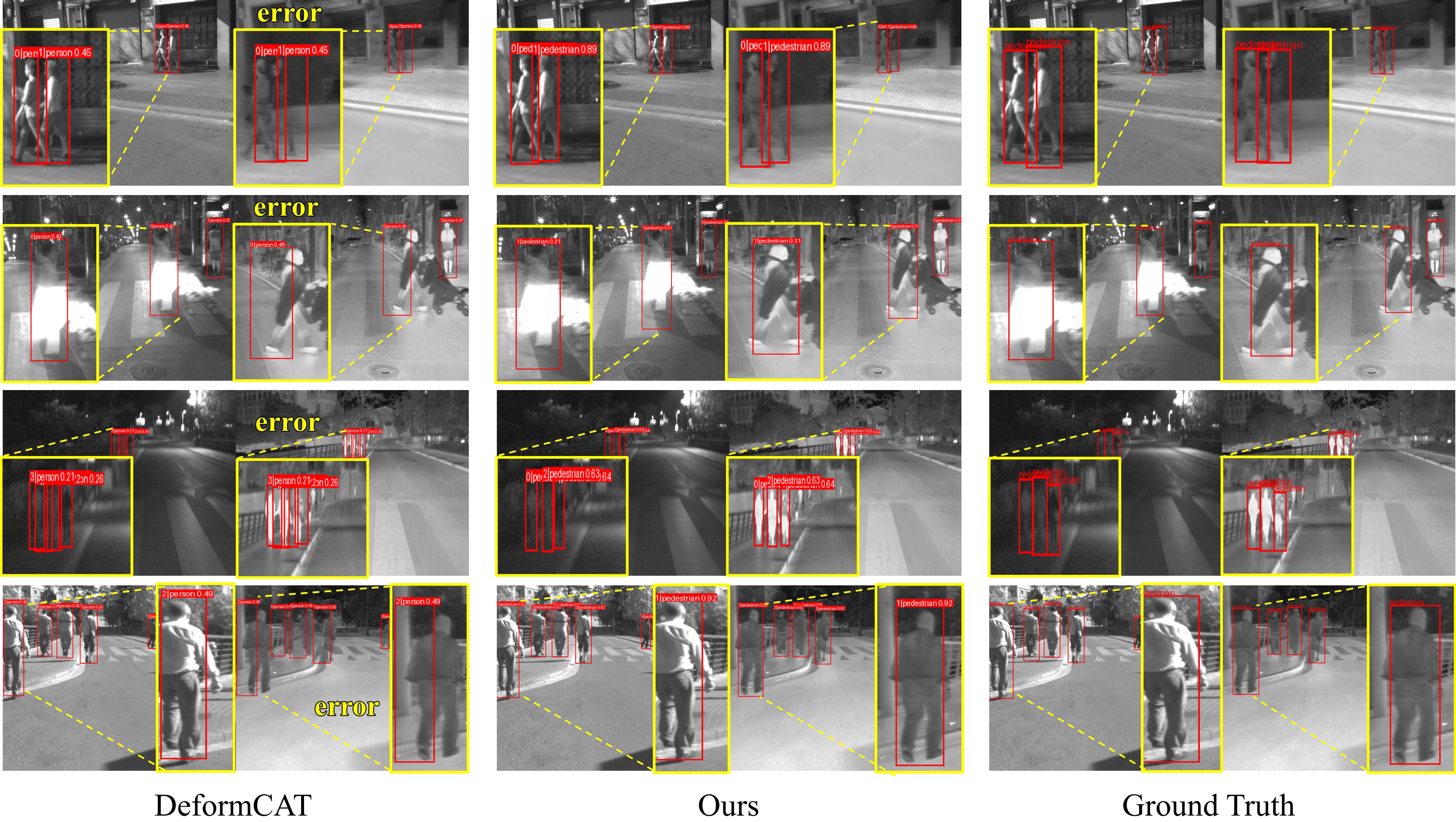}
\caption{Representative results of different methods on the CVC-14 dataset. For each method, the first column displays visible images, and the second column displays infrared images. The ``error'' denotes mistakes caused by detection offsets. The confidence threshold is set to 0.2 when visualizing these results.}
\label{fig:cvc-14}
\vspace{-1em}
\end{figure*}

\textbf{Qualitative comparison.} Fig.~\ref{fig:dronevehicle} and Fig.~\ref{fig:cvc-14} show some representative detection results of different methods on the DroneVehicle and CVC-14 datasets. It can be observed that CCLDet and DeformCAT only produce results for one modality, leading to large positional errors in the other. As shown in the yellow zoomed-in areas marked with ``error'' in the column, the detection boxes completely deviate from the objects themselves, which is unacceptable in practical applications. In contrast, our DOD-SA accurately predicts decoupled object positions in both infrared and visible modalities, even in low-light or highly misaligned scenarios. Furthermore, our model exhibits fewer classification errors in scenarios where other methods frequently make mistakes, as highlighted by the white circles. This is attributed to the fact that DOD-SA learns richer image features via the consistency regularization of the teacher-student network and receives accurate decoupled supervision during training, thereby achieving strong performance in both modalities.

\subsection{Ablation Studies}
We conducted extensive ablation studies on the DroneVehicle dataset to validate the effectiveness of our  method.

\textbf{Effect of PaST.} As shown in Table.~\ref{stage_results}, we analyze the necessity of each stage in PaST by ablating different training stages while guaranteeing full convergence for every stage. Comparing ours and Exp.~\MakeUppercase{\romannumeral 4} shows Stage 1 improves $m$AP50 by 0.67\%, as it helps the single-modality detector develop initial pseudo-label mining capability, ensuring better guidance for Stage 2. Exp.~\MakeUppercase{\romannumeral 3} vs. Exp.~\MakeUppercase{\romannumeral 4} verifies the importance of SM-Branch in guiding the training of DMD-Branch  in Stage 2. Finally, Exp.~\MakeUppercase{\romannumeral 5} vs. ours shows that Stage 3 boosts $m$AP50 by 0.86\% through teacher-student mutual learning, further refining the dual-modality branch.
\begin{table}[h]
\vspace{-1ex}
\centering
\caption{Ablation study of PaST.}
\label{stage_results}
\fontsize{8}{10}\selectfont
\begin{tabularx}{0.8\columnwidth}{ w{c}{0.5cm} | Y Y Y | w{c}{0.5cm} w{c}{0.5cm} }
\toprule 
Exp. & Stage 1 & Stage 2 & Stage 3 & IR & VIS \\
\midrule 
\MakeUppercase{\romannumeral 1} & \checkmark & &  & 75.75 & 48.70 \\
\MakeUppercase{\romannumeral 2} & & \checkmark &  & 79.44 & 77.68 \\
\MakeUppercase{\romannumeral 3} & & & \checkmark & 79.19 & 65.59 \\
\MakeUppercase{\romannumeral 4} & & \checkmark & \checkmark & 79.74 & 78.18 \\
\MakeUppercase{\romannumeral 5} & \checkmark & \checkmark &  & 79.55 & 77.77 \\
\midrule
\textbf{Ours} & \checkmark & \checkmark & \checkmark & \textbf{80.41} & \textbf{78.87} \\
\bottomrule
\end{tabularx}
\vspace{-1em}
\end{table}

\textbf{Effect of PLF.} We evaluate the impact of varying the threshold value $\tau$ in the Pseudo Label Filter (PLF). As shown in Table.~\ref{PLF}, the model performance degrades significantly without pseudo-label filtering (w/o PLF). The $m$AP50 varies with different thresholds, showing that a fixed $\tau$ cannot balance pseudo-label quantity and quality. Dynamically adjusting $\tau$ through distribution adaptation yields the best results, demonstrating PLF's importance for model training.
\begin{table}[h]
\centering
\caption{Ablation study of PLF.}
\label{PLF}
\fontsize{8}{10}\selectfont
\begin{tabularx}{0.8\columnwidth}{ w{c}{0.5cm} | Y | w{c}{0.5cm} w{c}{0.5cm} }
\toprule
Exp. & method & IR & VIS \\
\midrule
\MakeUppercase{\romannumeral 1} & w/o PLF    & 78.99 & 77.85 \\
\MakeUppercase{\romannumeral 2} & $\tau$ = 0.4 & 79.01 & 77.72 \\
\MakeUppercase{\romannumeral 3} & $\tau$ = 0.6 & 79.51 & 78.47 \\ 
\MakeUppercase{\romannumeral 4} & $\tau$ = 0.8 & 79.21 & 78.24 \\
\midrule
\textbf{Ours} & PLF & \textbf{80.41} & \textbf{78.87} \\
\bottomrule 
\end{tabularx}
\vspace{-2ex}
\end{table}

\textbf{Effect of SDLM.} Results are shown in Table~\ref{SDLM}. The ``w/o pseudo-labels'' setting implies a lack of training supervision for the visible modality. This causes the model to erroneously assume that there are no objects in this modality, reducing its VIS $m$AP50 to 12.63\%. Conversely, directly copying labels across modalities (w/o SDLM) degrades both modalities due to spatial misalignment. Our baseline, the IoU Matching Strategy, employs a label-pairing method similar to that used in the DroneVehicle dataset, which increases the $m$AP50 for both modalities. In contrast, SDLM introduces a search-region constraint for label matching, the $m$AP50 further improves by 0.86\% on IR and 1.35\% on RGB. These results confirm the effectiveness of our approach.
\begin{table}[t]
\centering
\caption{Ablation study of SDLM.}
\label{SDLM}
\fontsize{8}{10.5}\selectfont
\begin{tabularx}{0.8\columnwidth}{ w{c}{0.5cm} | Y | w{c}{0.5cm} w{c}{0.5cm} }
\toprule
Exp. & Method & IR & VIS \\
\midrule 
\MakeUppercase{\romannumeral 1}&w/o SDLM&79.14 &77.47  \\
\MakeUppercase{\romannumeral 2}&w/o pseudo-labels & 78.50  & 12.63 \\
\MakeUppercase{\romannumeral 3}&IOU Match Strategy  & 79.55  & 77.52 \\
\midrule
\textbf{Ours} &SDLM & \textbf{80.41} & \textbf{78.87} \\
\bottomrule 
\end{tabularx}
\end{table}

\textbf{Effect of DLC.} We perform an ablation study to evaluate the impact of Dynamic Label Correction (DLC). As shown in Table.~\ref{DLC}, using only matched label pairs (w/o DLC) results in 79.18\% $m$AP50 on IR. Introducing unmatched labels (w/o DU) improves performance, highlighting their importance. Cross‐modality BBox Correction (CBC) \cite{zhang2025mitigating} uses a progressive center correction strategy to adjust shifted ground‐truth boxes in the unlabeled modality, which may introduce label ambiguity because the smoothed labels can be misplaced, thus misleading the model during training. In contrast, our method dynamically updates the label bag to mitigate this issue, yielding performance improvements of 0.76\% on IR and 0.75\% on VIS compared to CBC.
\begin{table}[t]
\vspace{-2ex}
\centering
\caption{Ablation study of DLC.}
\label{DLC}
\fontsize{8}{10}\selectfont
\begin{tabularx}{0.8\columnwidth}{ w{c}{0.5cm} | Y | w{c}{0.5cm} w{c}{0.5cm} }
\toprule
Exp. & Method & IR & VIS \\
\midrule
\MakeUppercase{\romannumeral 1}&w/o DLC& 79.18  &77.78  \\
\MakeUppercase{\romannumeral 2}&w/o DU& 79.27 &78.57\\
\MakeUppercase{\romannumeral 3}&CBC& 79.65  & 78.12 \\
\midrule
\textbf{Ours} & DLC & \textbf{80.41} & \textbf{78.87} \\
\bottomrule
\end{tabularx}
\end{table}

\begin{figure}[t]
	\centering
	\includegraphics[width=1.0\linewidth]{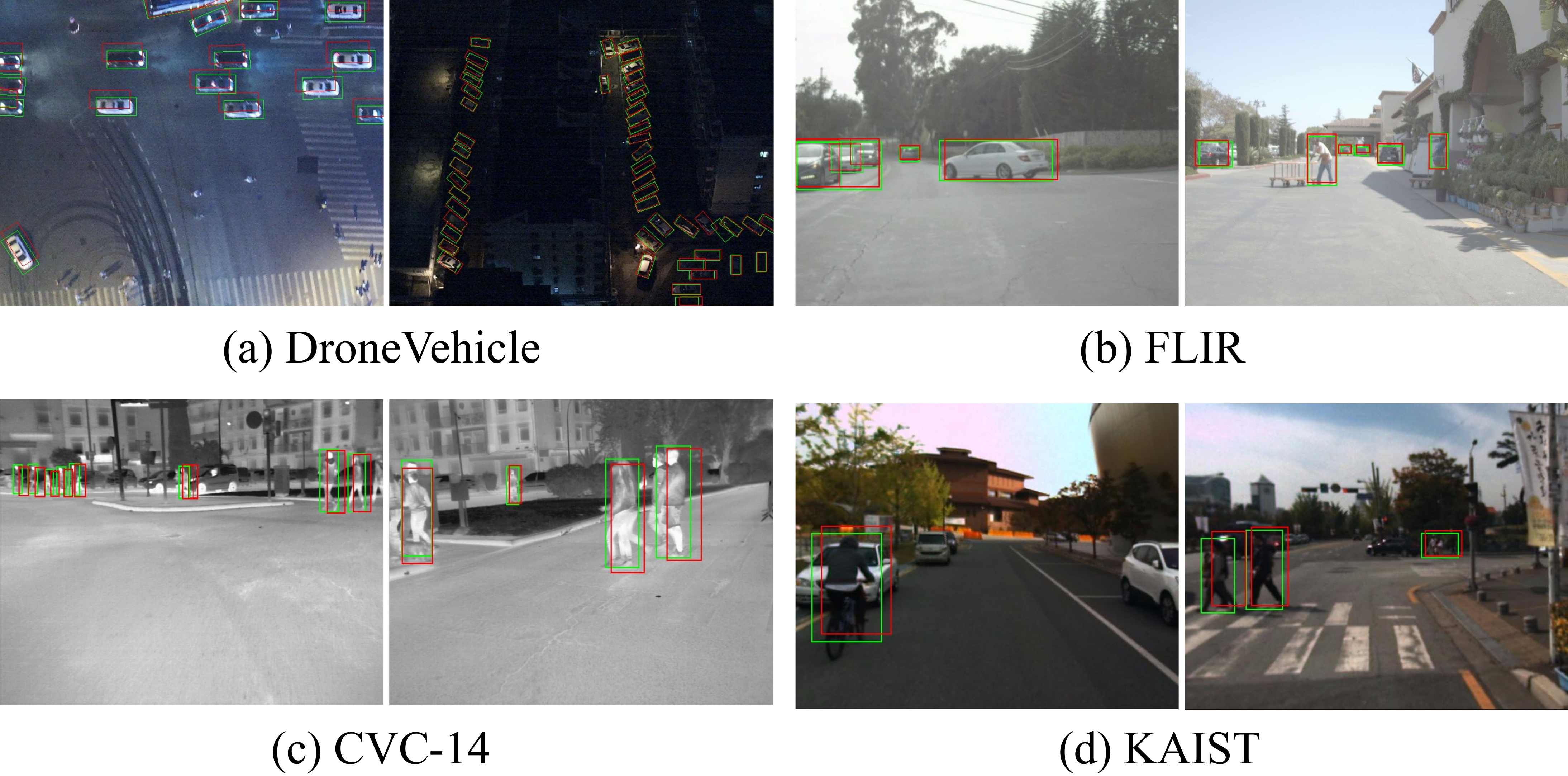}
	\caption{Visualization of pseudo-labels on four public datasets. Red boxes represent the ground truth copied from the labeled modality to the unlabeled modality, while green boxes represent the refined pseudo-labels in the unlabeled modality.}
	\label{fig:shift}
\end{figure}

\textbf{Effectiveness of Pseudo-Label Mining.} To illustrate the efficacy of pseudo-label correction on the unlabeled modality, we visualize the pseudo-labels mined by our model across the four datasets. These pseudo-labels are extracted for visualization at the end of the third training stage. Fig.~\ref{fig:shift}(a) illustrates that our model effectively corrects offsets induced by high-speed object motion. Even when objects are stationary, misalignment may still arise due to varying shooting angles and drone movement; nevertheless, our method successfully rectifies this issue. Furthermore, Fig.~\ref{fig:shift}(b)-(d) demonstrate that our approach maintains robustness across a variety of scenarios.

\subsection{Robustness to Position Shift}
\begin{figure}[t]
	\centering
	\vspace*{-\topskip}
	\includegraphics[width=0.8\linewidth]{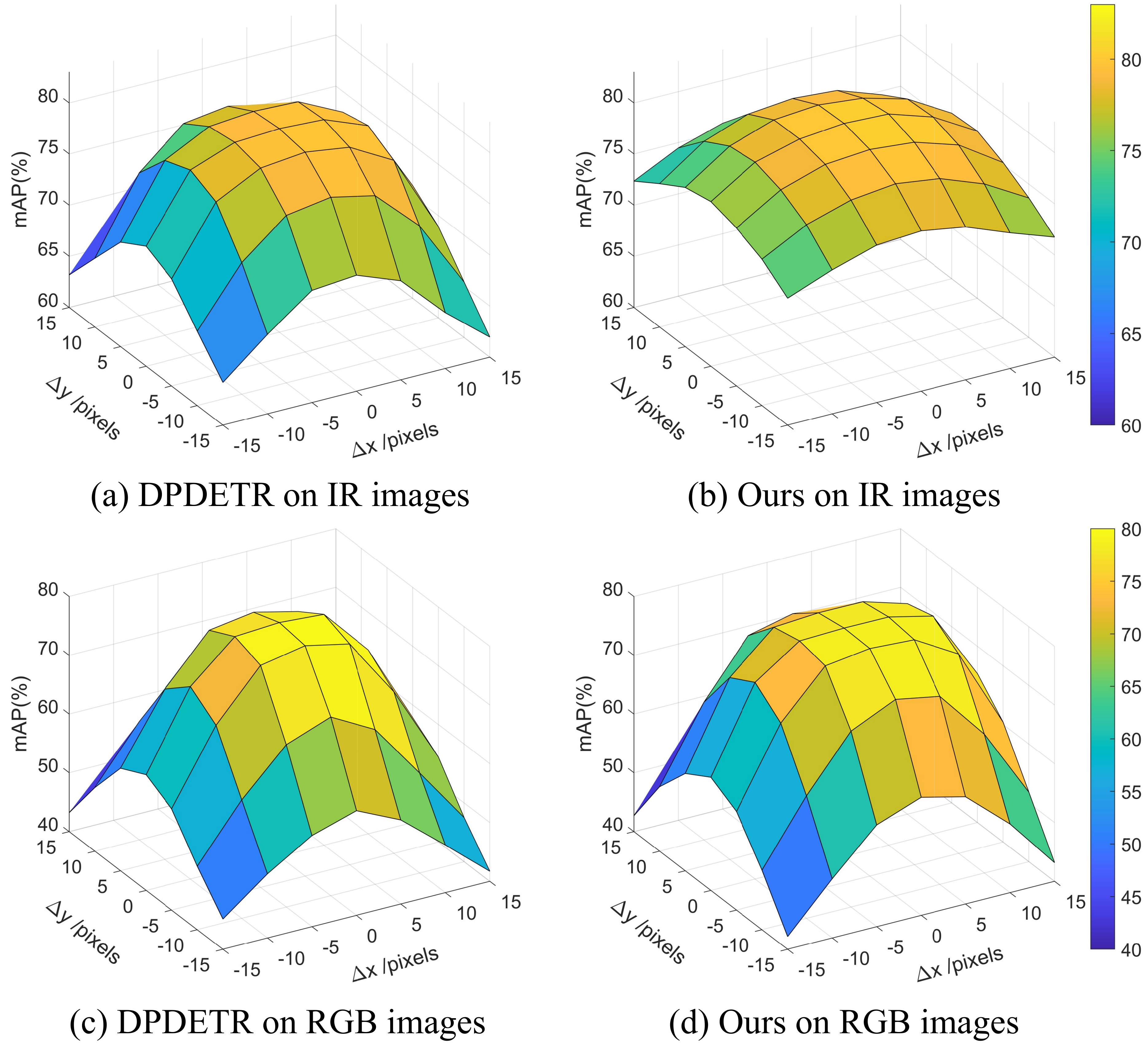}
	\caption{Performance comparison under various positional shifts. Note that the labeled modality is infrared.}
	\label{fig:robust_exp} 
	\vspace{-1.5em} 
\end{figure}

To evaluate the robustness of our method against the modality misalignment, we introduce positional offsets to the test image set. We keep the IR images fixed while applying offsets along both the x-axis and y-axis to the visible images. The pixel value variations are defined within the range \(\{(\Delta x, \Delta y) \mid \Delta x, \Delta y \in [-15, 15]; \Delta x, \Delta y \in \mathbb{Z}\}\). As shown in Fig.~\ref{fig:robust_exp}, the performance of our method on the IR test set is more stable under these shifts compared to the baseline. For the VIS test set, both models perform similarly, but our method is slightly more stable under small shifts. This indicates that our method enhances the model's understanding of the correspondences between each pair of objects across the two modalities and makes it more robust to positional misalignment.

\section{Conclusion}
In this paper, we propose DOD-SA, a novel infrared-visible decoupled object detection framework with single-modality annotations. Our Collaborative Single- and Dual-Modality Teacher-Student Network (CoSD-TSNet) leverages the single-modality branch (SM-Branch) to guide the training of the dual-modality decoupled branch (DMD-Branch) via a Progressive and Self-Tuning Training Strategy (PaST), which addresses the modality gap and improves the quality of pseudo-labels. Additionally, we introduce a Pseudo Label Assigner (PLA) to match the ground-truth from the labeled modality with mined pseudo-labels from the unlabeled modality. Experiments on four public datasets demonstrate the superiority of DOD-SA over SOTA methods. Ablation studies validate the necessity of each component, and our method exhibits strong robustness to positional shifts across position shift experiments.
\bibliographystyle{IEEEtran}  
\bibliography{main}           

\end{document}